\documentclass{article}
\usepackage[preprint]{neurips_2022}

\usepackage[utf8]{inputenc} 
\usepackage[T1]{fontenc}    
\usepackage{hyperref}       
\usepackage{url}            
\usepackage{booktabs}       
\usepackage{amsfonts}       
\usepackage{nicefrac}       
\usepackage{microtype}      
\usepackage[usenames,dvipsnames]{xcolor}
\usepackage{amsmath, amsthm, amssymb, bm, bbm}
\usepackage{multirow}
\usepackage{subfigure}
\usepackage{graphicx}
\usepackage{caption}
\usepackage{soul}
\usepackage{wrapfig}
\usepackage{algorithm}
\usepackage{algpseudocode}
\usepackage{mwe}
\usepackage{caption} 
\DeclareUnicodeCharacter{221A}{${\sqrt{ }}$}

\begin{document}
\title{A Study on the Evaluation  of Generative Models}
%

\author{%
  Eyal Betzalel
  \\\texttt{eyal.betzalel@biu.ac.il} 
  \And
  Coby Penso
  \\\texttt{coby.penso@biu.ac.il}
  \AND
  Aviv Navon
  \\\texttt{aviv.navon@biu.ac.il}
  \And
  Ethan Fetaya
  \\\texttt{ethan.fetaya@biu.ac.il}
  \AND {\normalfont Bar-Ilan University, Israel}
}

\date{Affiliation}

\maketitle
\begin{abstract}
    Implicit generative models, which do not return likelihood values, such as generative adversarial networks and diffusion models, have become prevalent in recent years.
    While it’s true that these models have shown remarkable results, evaluating their performance is challenging. This issue is of vital importance to push research forward and identify meaningful gains from random noise. 
    Currently, heuristic metrics such as the Inception score (IS) and Fr\'{e}chet Inception Distance (FID) are the most common evaluation metrics, but what they measure is not entirely clear. Additionally, there are questions regarding how meaningful their score actually is. In this work, we study the evaluation metrics of generative models by generating a high-quality synthetic dataset on which we can estimate classical metrics for comparison. Our study shows that while FID and IS do correlate to several f-divergences, their ranking of close models can vary considerably making them problematic when used for fain-grained comparison. We further used this experimental setting to study which evaluation metric best correlates with our probabilistic metrics. Lastly, we look into the base features used for metrics such as FID.
\end{abstract}

\section{Introduction}
\label{Introduction}
Implicit generative models  such as Generative Adversarial Networks (GANs) \cite{GAN} and diffusion models \cite{diffusion} have made significant progress in recent years, such as developing the capability to generate high-quality images \cite{StyleGANv2,dalle2} and audio \cite{HiFiGAN}.  Despite these successes, evaluation is still a major challenge for implicit models that do not predict likelihood values. While significant improvement can easily be observed visually, at least for images, an empirical measure is required as an objective criterion and for comparison between relatively similar models. Moreover, devising objective criteria is vital for development, where one must choose between several design choices, hyperparameters, etc.
The most common practice is to use metrics such as Inception score (IS) \cite{IS} and Fr\'{e}chet Inception Distance (FID) \cite{heusel2018gans} that are based on features and scores computed using a network pre-trained on the ImageNet \cite{5206848} dataset. While these proved to be valuable tools, they have some key limitations: (i) It is unclear how they relate to any classical metrics on probabilistic spaces. (ii) These metrics are based on features and classification scores trained on a certain dataset and image size, and it is not clear how well they transfer to other image types, e.g. human faces, and image sizes,  (iii) The scores can heavily depend on particular implementation details \cite{note_on_IS,buggy_FID}. 

Another evaluation tool is querying humans. One can ask multiple human annotators to classify an image as real or fake or to state which of two images they prefer. While this metric directly measures what we commonly care about in most applications, it requires a costly and time-consuming evaluation phase. Another issue with this metric is that it does not measure diversity, as returning a single good output can get a good score.  

To better understand these evaluation metrics and deep generative models in general, we create a high-quality synthetic dataset, using the powerful Image-GPT model \cite{image-gpt}. This is a complex synthetic data distribution that we can sample from and compute exact likelihood values.  As this data distribution is trained on natural images from the ImageNet dataset using a strong model, we expect the findings on it to be relevant to models trained on real images. 
The dataset provides a solid and useful test-bed for developing and experimenting with generative models.
We will make our dataset public \footnote{github.com/eyalbetzalel/notimagenet32}  for further research. 


Using this test-bed we train various likelihood models and evaluate their KL-divergence and reverse KL-divergence. 
We then compare the scores and ranking given by these divergences to empirical metrics such as FID and observe that while the empirical metrics correlate nicely to these divergences, they are much more volatile and thus might not be well-suited for fine-grained comparison. 
  
Finally, we investigated the use of the Inception network \cite{SzegedyLJSRAEVR15} for feature extraction on FID, specifically for image datasets that are different from the ImageNet on which it was trained. This is important as FID is commonly used to compare models on datasets such as CelebA (human faces) \cite{liu2015faceattributes} and LSUN \cite{https://doi.org/10.48550/arxiv.1506.03365} bedrooms that are quite distinct from ImageNet. Specifically, we investigate the Gaussianity assumption that lies in the base of the FID metric, compared to features returned by CLIP \cite{DBLP:journals/corr/abs-2103-00020} which was trained on a wider variety of images. We show both quantitatively and qualitatively that the CLIP features are better suited than the Inception features on non-ImageNet datasets.



\section{Background}\label{sec:background}
Given the popularity of GANs and other implicit generative models, many heuristic evaluation metrics have been proposed in recent years. We give a quick overview of the most common metrics and probabilistic KL-divergences. 

\subsection{$KL$-Divergence}
One common measure of difference between probability distributions is the the Kullback–Leibler (KL) divergence $KL(p||q)=\mathbb{E}_{x\sim p}\left[\log\left(\frac{p(x)}{q(x)}\right)\right]$, noting that it is not symmetric. We refer to $KL(p_{data}||p_{model})$ as the KL divergence and $KL(p_{model}||p_{data})$ the Reverse KL (RKL) divergence.
 The KL is commonly used even when $p_{data}$ is unknown, which is the standard case, as it can still be estimated from samples up to a constant. Also, standard maximum likelihood optimization is equivalent to minimizing the empirical KL divergence. It is important to note that the KL divergence is biased towards ``inclusive" models (where the model ``covers" all high likelihood areas of the data distribution). The RKL has a bias toward ``exclusive" models (where the model does not cover low likelihood areas of the data distribution). While an exclusive bias might be more appropriate in some applications, such as out-of-distribution detection, we cannot optimize it directly without access to $p_{data}$.
\subsection{Inception Score}
Inception Score (IS) is a metric for evaluating the quality of image generative models based on InceptionV3 Network pre-trained on ImageNet. It calculates:
$$ IS = \exp\left( E_{x\sim{p_{G}}}[KL(p_\theta(y|x) || p_{\theta}(y)]\right)$$
where $x\sim p_G$ is a generated image, $p_\theta(y|x)$ is the conditional class distribution computed via the inception network, and $p_{\theta}(y) = \int_{x} p_\theta(y|x)p_{G}(x)dx$ is the marginal class distribution.\\
The two desired qualities that this metric aims to capture are:
(i) The generative model should output a diverse set of images from all the different classes in ImageNet, i.e $p_{\theta}(y)$ should be uniform (ii) The images generated should contain clear objects so the predicted probabilities $p_\theta(y|x)$  should be close to a one-hot vector and have low entropy.
When both of this qualities are satisfied then the KL distance between $p_\theta(y)$ and $p_\theta(y|x)$ is maximized. Therefore the higher the IS is, the better. 
 
\subsection{Fréchet Inception Distance}
The FID metric is based on the assumption that the features computed by a pre-trained Inception network, for both real and generated images, have a Gaussian distribution. We can then use known metrics for Gaussians as our distance metric. 
Specifically, FID uses the Fréchet distance between two multivariate Gaussians which has a closed-form formula. For both real and generated images we fit Gaussian distributions to the features extracted by the inception network at the pool3 layer and compute 
$$FID = ||\mu_r - \mu_g||^2 + Tr(\Sigma_r +\Sigma_g -2(\Sigma_r \Sigma_g)^{1/2})$$
where $\mathcal{N}(\mu_r,\Sigma_r)$ and $\mathcal{N}(\mu_g,\Sigma_g)$ are the Gaussian fitted to the real and generated data respectively. The quality of this metric depends on the features returned by the inception net, how informative are they about the image quality, and how reasonable is the Gaussian assumption about them. 



\subsection{Kernel Inception Distance}
The Kernel Inception Distance (KID)  \cite{binkowski2018demystifying} aims to improve on FID by relaxing the Gaussian assumption. KID measures the squared Maximum Mean Discrepancy (MMD) between the Inception representations of the real and generated samples using a polynomial kernel. This is a non-parametric test so it does not have the strict Gaussian assumption, only assuming that the kernel is a good similarity measure. It also requires fewer samples as we do not need to fit the quadratic covariance matrix.


\subsection{FID$_\infty$ \& IS$_\infty$}
In \cite{chong2020effectively} the authors show that the FID and IS metrics are biased when they are estimated from samples and that this bias depends on the model. As the bias is model-dependent, it can skew the comparison between different models. The authors then propose unbiased version of FID and IS named FID$_\infty$ / IS$_\infty$.



\subsection{Clean FID}  

As the input to the Inception network is fix-sized, generated images of different sizes need to be resized to fit the network's desired input dimension. The work in \cite{parmar2022aliased} investigates the effect of this resizing on the FID score, as the resizing can cause aliasing artifacts. The lack of consistency in the processing method can lead to different FID scores, regardless of the generative model capabilities. They introduce a unified process that has the best performance in terms of image processing quality and provide a public framework for evaluation. 

\subsection{Related work}

Here we add additional related work besides all the previously covered metrics. \cite{Bond_Taylor_2021} performed a comparative review of deep generative models. For the scope of this article, it is important to understand the difference between likelihood-based methods such as VAE and AR and non-likelihood based such as GANs and diffusion models. \cite{borji} presents a comprehensive survey of generative model estimation methods. \cite{noteonis} first pointed out issues in IS. \cite{normality} inspect the distribution of the Inception latent feature and suggest a more accurate model for evaluation purposes. \cite{empirical} perform an empirical study on an older class of evaluation metrics of GANs and mention that KID outperforms FID and IS.  


\section{Synthetic dataset as a benchmark}

\begin{figure*}[!h]
    \centering
    \captionsetup{justification=centering}

    \includegraphics[scale=0.6]{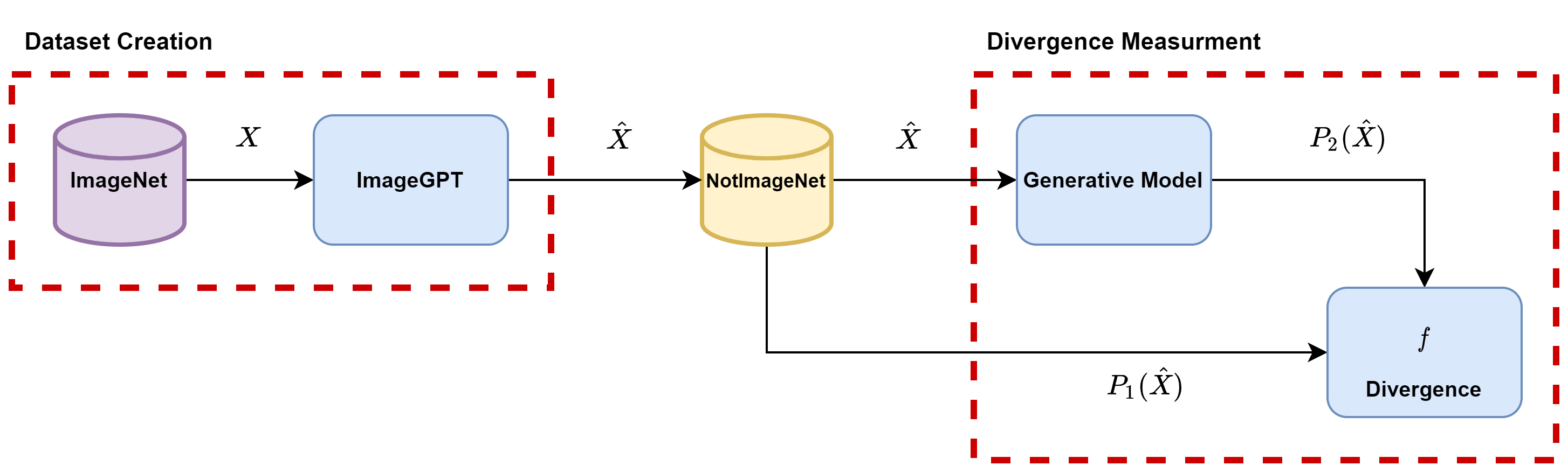}
    \caption {Illustration: $X$ are ImageNet images, $\hat{X}$ are synthetic images that sampled from image-GPT, $P_1(\hat{X})$ is ground truth likelihood from image-GPT for synthetic images and $P_2(\hat{X})$ is likelihood estimation of $P_1(\hat{X})$, calculated by the evaluated model, in this case, PixelSnail.}
\end{figure*}
We created an auxiliary dataset of 100K images by sampling images from the  Image-GPT model that has been trained on ImageNet32, the ImageNet dataset that was resized to  $32\times 32$. We split the data set into a training set (70K images) and a test set (30K images). As this is a synthetic version of ImageNet32 we name our dataset \textit{NotImageNet32}. 

We note that Image-GPT clusters the RGB values of each pixel into 512 clusters and predicts these cluster indexes. This means that instead of each pixel corresponding to an element of $\{0,...,255\}^3$ it belongs to $\{0,...,511\}$. We can map these cluster values back to RGB, as was done in Image-GPT, for visualization. 

\begin{figure*}[!h]
    \centering
    \captionsetup{justification=centering}
    \includegraphics[scale=0.3]{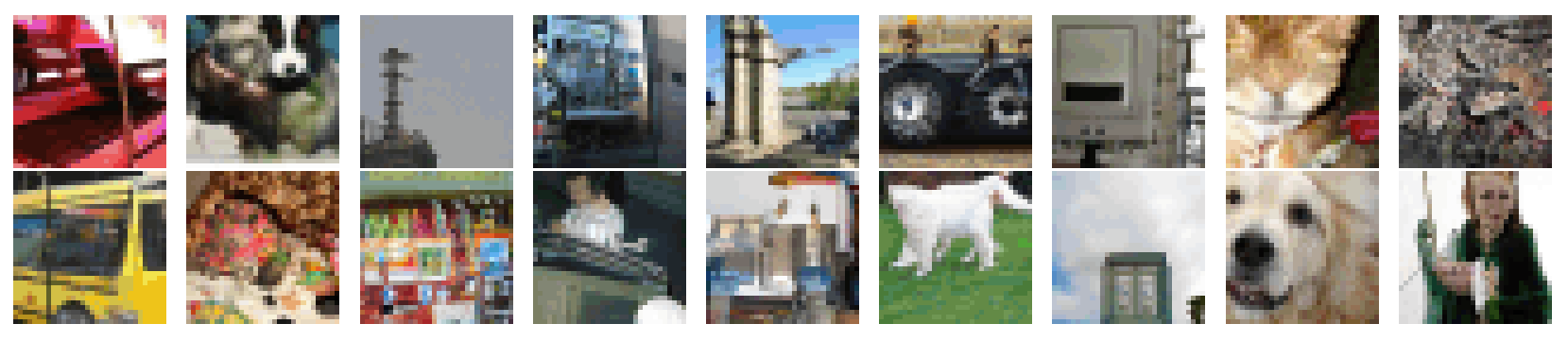}
    \caption {Examples of photos that generated by image-GPT. Each photo explicit likelihood can be measured.}
    \label{fig:imageGPT}
\end{figure*}

To evaluate and understand generative model metrics we train a set of models on this dataset. One set of models is based on the PixelSnail model \cite{chen2017pixelsnail} which uses an autoregressive generative technique, and the other set is based on VD-VAE \cite{child2021deep} which is VAE \cite{kingma2014autoencoding} based (we used IWAE \cite{burda2016importance} to reduce the gap between the ELBO and the actual likelihood). We note that all models were adjusted to our dataset and output the clustered index instead of RGB values. Supplementary details on the models architecture in this experiment can be found in the appendix section.

To produce a diverse set of models with varying degrees of quality, each set was trained several times with different model sizes. We save a model for comparison after every five epochs of training. As a result, the models we compare are a mix of strong and weak models.
After the training procedure, we can compute for each image in the test set its likelihood score (or the IWAE bound) for each model.

We then measure the difference between the original distribution $p_{data}(x)$ and the approximated distribution $p_{GM}(x)$ by using Monte-Carlo approximation of two divergence function: Kullback–Leibler (KL) $KL(p_{data}||p_{GM})$ and Reverse KL (RKL) $KL(p_{GM}||p_{data})$.
As these divergences measure complementary aspects, one inclusive and one exclusive, we believe that this gives us a well-rounded view of the generative model behavior.

A limitation of this test-bad is that it can be applied only to likelihood-based models, so implicit models like GAN are not able to take advantage of it. Additionally, although KL and RKL are complementary, they are unable to capture all the complexities of a generative model.  




\section{Comparison between evaluation metrics}\label{sec:expiriment}
\subsection{Volatility}

\begin{figure*}[!h]
    \centering
    \captionsetup{justification=centering}
    \includegraphics[scale=0.3]{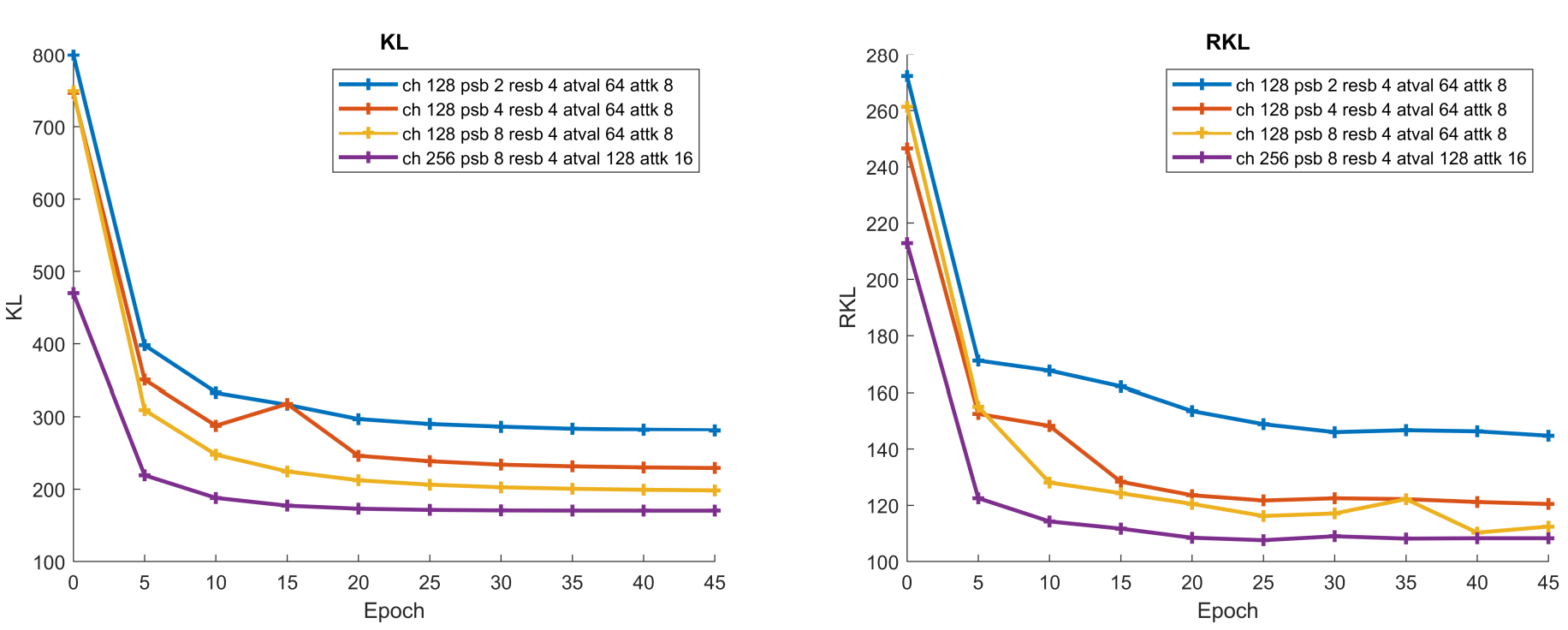}
    \caption {Test KL and RKL of PixelSnail models along training.}
    \label{fig:pixelsnail_training1}
\end{figure*}


\begin{figure*}[!h]
    \centering
    \captionsetup{justification=centering}
    \includegraphics[scale=0.3]{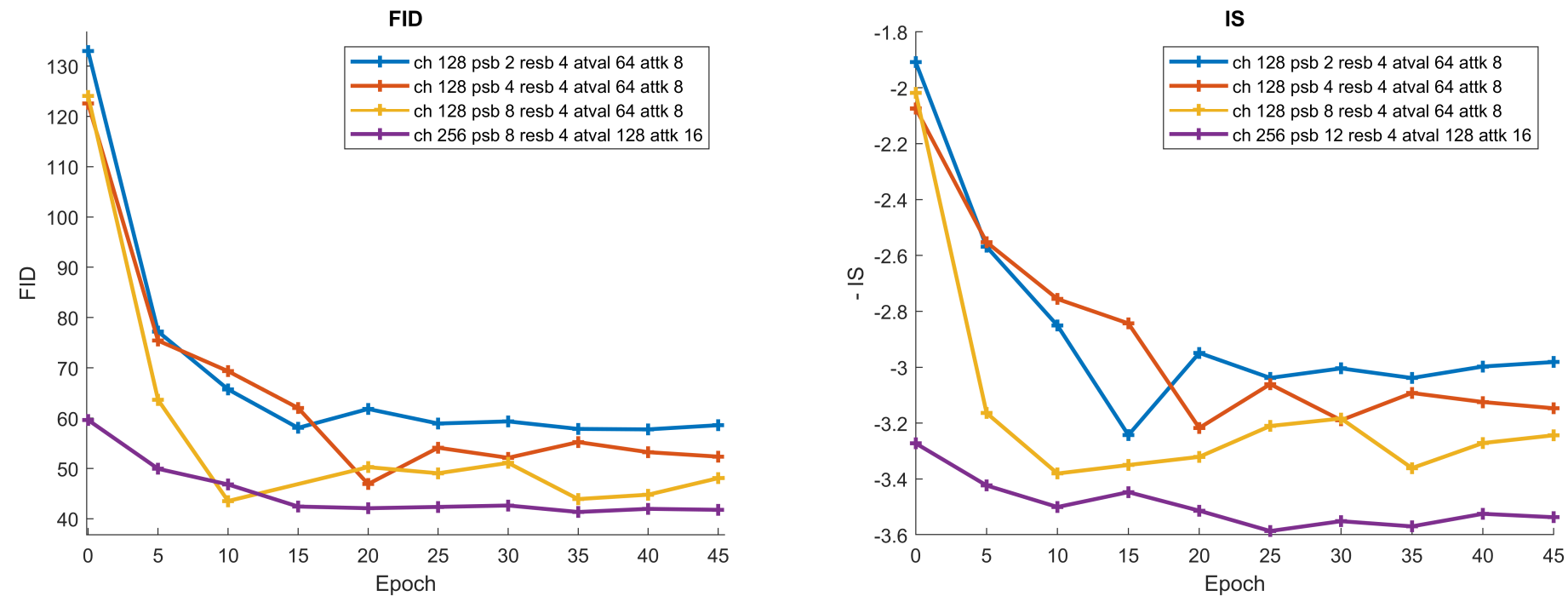}
    \caption {Test FID and negative IS of PixelSnail models along training. We plot the negative Inception Score so lower is better for all metrics. Details on the hyperparameters summerized in the legend are in the appendix.}
    \label{fig:pixelsnail_training2}
\end{figure*}


We first train four PixelSnail variants on our NotImageNet32 dataset and plot the KL, RKL, FID, and IS along with the training in Fig. \ref{fig:pixelsnail_training1} and \ref{fig:pixelsnail_training2}.
It can easily be seen that after 15-20 epochs both KL and RKL change slowly, but the FID and IS are much more volatile. Each dot in the graph represents a score that has been measured on a different epoch on a different model. One can see from this figure that as we increase the model capacity, the KL score improves. Interestingly, the KL and RKL have a high agreement even if they penalize very different mistakes in the model. In stark contrast, we see that the FID, and especially IS, are much more volatile and can give very different scores to models that have very similar KL and RKL scores.

\begin{figure}[!h]\label{fig:scatters}.
\captionsetup{justification=centering}
\centering
\includegraphics[width = \textwidth]{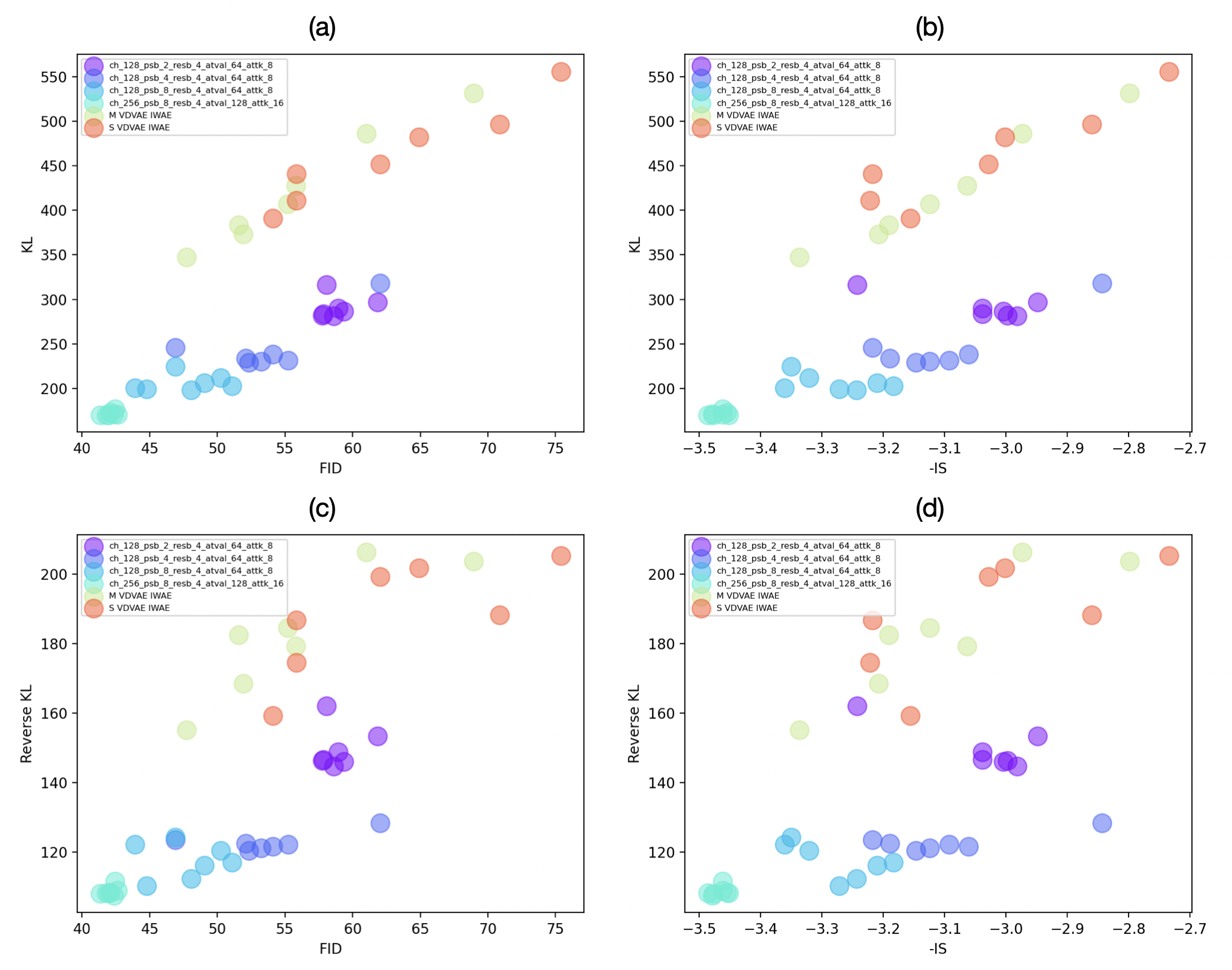}
\caption{Evaluation metrics along the training of four pixelsnail and two VD-VAE models of varying sizes. We plot the negative Inception Score so lower is better for all metrics.}
\label{fig:scatters}
\end{figure}
To get another perspective, we plot in Fig. \ref{fig:scatters} the FID and negative IS vs. KL and RKL. We observe a high correlation between FID/IS and KL and a weaker correlation between these metrics and the RKL. IS and FID are also seem ill-suited for fine-grained comparisons between models. For high-quality models, e.g., light-blue dots in Fig. \ref{fig:scatters}, one can get a significant change in FID/IS without a significant change to KL/RKL. We add zoomed-in versions of this plot to the appendix for greater clarity.



\subsection{Ranking correlation}

To better quantitatively assess our previous observations, we compare how the metrics differ in their ranking of the various trained models. This is of great importance, as comparing different models is the primary goal of these metrics. To compare the ranking we compute Kendall's $\tau$ ranking correlation (Tab. \ref{tab:kandell}). We perform the correlation analysis for models that were trained for 15 - 45 epochs and ignore the first iterations of the training procedure. This is done to focus more on the fine-grained comparisons.

\newcommand{\bftab}{\fontseries{b}\selectfont}
\begin{table}[hbt!]\label{tab:kandell}
    \centering
    \caption{Kandell's $\tau$ Correlation}
    \begin{tabular}{c c c c c c c c c}
    \toprule
       & KL & RKL & FID & IS & IS$_\infty$ & KID & FID$_\infty$ & Clean FID \\ \midrule
        KL & 1 & \bftab {0.8895} & 0.7027 & 0.5889 & 0.4681 & 0.7770 & 0.8095 & 0.7909 \\ 
        RKL & \bftab {0.8895} & 1 & 0.6337 & 0.5244 & 0.4314 & 0.7105 & 0.7267 & 0.7198 \\ \midrule
        FID & 0.7027 & 0.6337 & 1 & 0.7979 & 0.7189 & 0.8513 & 0.8002 & 0.8699 \\
        IS & 0.5889 & 0.5244 & 0.7979 & 1 & 0.8281 & 0.7329 & 0.6818 & 0.7236 \\ 
        IS$_\infty$ & 0.4681 & 0.4314 & 0.7189 & 0.8281 & 1 & 0.6167 & 0.5749 & 0.6074 \\ 
        KID & 0.7770 & 0.7105 & 0.8513 & 0.7329 & 0.6167 & 1 & 0.8606 & 0.9675 \\ 
        FID$_\infty$ & 0.8095 & 0.7267 & 0.8002 & 0.6818 & 0.5749 & 0.8606 & 1 & 0.8746 \\
        Clean FID & 0.7909 & 0.7198 & 0.8699 & 0.7236 & 0.6074 & 0.9675 & 0.8746 & 1 \\
        \bottomrule
    \end{tabular}
    \label{tab:kandell}
\end{table}
The highest score in both ranking correlation methods is between KL and Reverse KL with 0.889 Kendall's $\tau$. This may be counterintuitive since these two methods measure different characteristics of the data. Confirming our previous observation, the FID and IS ranking scores are low, with FID outperforming IS. However, the extensions of FID do achieve better scores. 

Another observation is the relatively low correlation between many of the different rankings. All of the Inception ranking correlation, except one  (KID and Clean FID), indicates that one can get significantly different rankings by using a different metric. 

Among the Inception-based metrics, FID$_{\infty}$ has the highest correlation with KL and RKL which indicates that it is a more reliable metric than the other. IS/IS$_\infty$ has the lowest ranking correlation between all other models.

\section{Is Inception all we need?}
Most common metrics for generative models are based on features computed by a pre-trained network. The standard features are the output of an Inception network trained on the ImageNet classification task. The underlying assumption behind FID and its extensions is that these features are representative of the quality of the image and that they follow a Gaussian distribution. 

As these metrics are used to evaluate generative models on various domains, e.g., faces, pets, bedrooms, etc., that are distinct from the ImageNet dataset on which the Inception network was trained, it raises the question: Are the features returned by the Inception network the right choice for comparing generative models on various datasets?

In the next section, we evaluate the Inception features qualitatively and quantitatively, and compare them to the features computed by the CLIP (Contrastive Language-Image Pre-Training) network. CLIP is a neural network trained on the task of matching images to captions. It was trained on 400M images from a wide variety of domains, and was shown in multiple works to give strong representations that are useful for generating images \cite{gal2021stylegannada,Galatolo_2021,https://doi.org/10.48550/arxiv.2111.13792}. We hypothesize that since CLIP was trained on multiple domains and using full image captions, its features would be better suited for comparing generative models.

\subsection{Qualitative analysis of the latent representation}
FID and its extensions are based on the assumption that the distribution over the latent representation is Gaussian. Here we evaluate how this Gaussian assumption holds. To do this we fit a Gaussian to the real data using each representation and look at the generated images that get the best/worst likelihood according to this Gaussian. In detail we: 

\begin{enumerate}
  \item Sample 10K images from a generative model.
  \item Randomly select 20K images from the original data set used to train the generative model and compute their feature vectors with the Inception network and the CLIP network.
  \item For each of these representations, fit a Gaussian model.
  \item Calculate the probability of each of the synthetic samples belonging to the corresponded Gaussian model and rank them by their score.
\end{enumerate}

\begin{figure}[h]
    \centering
    \begin{minipage}{.5\textwidth}
    \centering
    \begin{center}
    \begin{subfigure}
        \centering
        \includegraphics[scale=0.3]{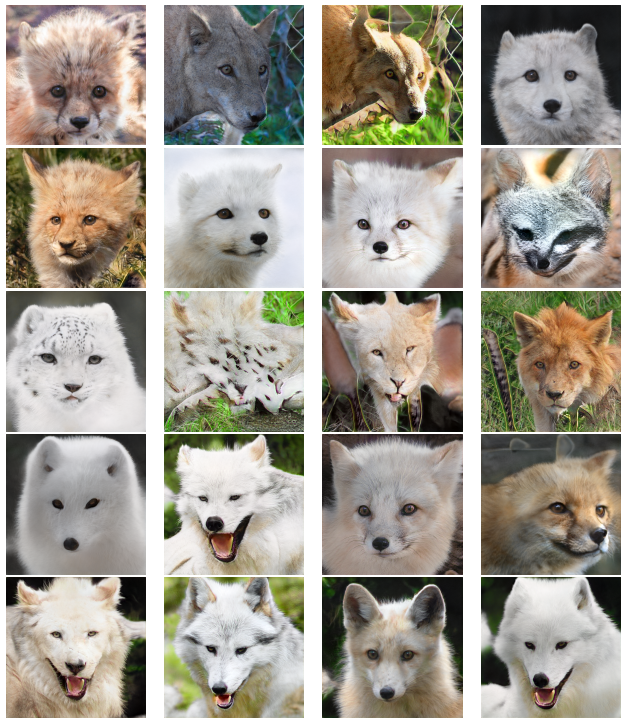}
        \caption{Inception - low probability (Wild).}
        \label{fig:low_inception_wild}
     \end{subfigure}
     \end{center}
     \end{minipage}%
\hfill
    \begin{minipage}{.5\textwidth}
    \centering
    \begin{center}
    \begin{subfigure}
        \centering
        \includegraphics[scale=0.3]{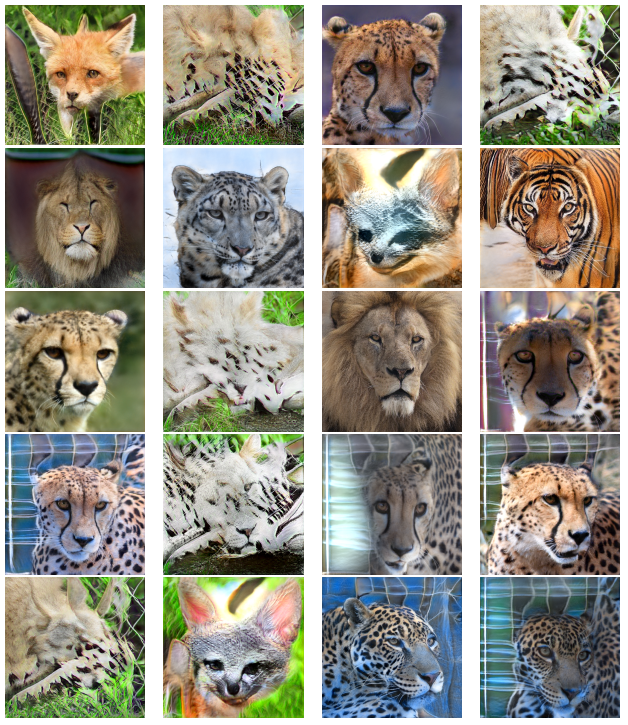}
        \caption{CLIP - low probability (Wild).}
        \label{fig:low_clip_wild}
     \end{subfigure}
     \end{center}
     \end{minipage}

\end{figure}

In Fig. \ref{fig:low_inception_wild} and \ref{fig:low_clip_wild} we show the images that got the lowest probability rank on the AFHQ dataset  \cite{https://doi.org/10.48550/arxiv.1912.01865} with the wild class. We used a pre-trained StyleGAN2-ADA \cite{https://doi.org/10.48550/arxiv.2006.06676} as our generative model. In the appendix, we provide additional examples with different generative models and datasets.


Since those images got the lowest probability rank among 10K images, our Gaussian model classifies them as outliers. If matching the Gaussian on these features is a good metric, then these low-probability images should correspond to low-quality generated images. 
As one can see, many of the CLIP low-probability images look indeed like anomalies while most of the Inception anomalies look like valid images from Wild dataset. While this is only a small number of images, we see similar behavior across various non-ImageNet datasets and generative models. These results indicate that CLIP features are better suited for comparison between generative models.


\subsection{Normality test for latent representation}
We augment our previous qualitative assessment by testing how well our samples follow a Gaussian distribution on our features. To utilize the readily available normality tests on 1D data, we linearly project our data randomly to one dimension and use these tests on multiple projections. This is valid, as a linear mapping of a multivariate normal also follows a normal distribution. Specifically, we: 

 \begin{enumerate}
  \item Propagate dataset with $N$ pictures via Inception and CLIP and save the latent vectors of the images. $A\in\mathbb{R}^{N \times d}$ is the result matrix, where $d$ is the latent representation dimension (2048 for Inception and 512 for CLIP).      
  \item Generate  $\mathbf{x}\in\mathbb{R}^d$ unit vector in a uniformly random direction.
  \item Calculate $ \mathbf{z} = A\mathbf{x}\in\mathbb{R}^N$, The projection of $A$ on random direction $\mathbf{x}$.
  \item Run the D'Agostino's K-squared normality test \cite{normaltest} and calculate $p$ value under the null hypothesis that the data were drawn from a Gaussian distribution.
  
  \item Repeat the process for $T=1000$ times for different randomized unit vectors and calculate the mean of $p$ value. 

  \end{enumerate}

\begin{table}[!ht]
    \centering
    \caption{Mean $p$ value results of normality test}
    \begin{tabular}{c c c c c c}
    \toprule
        ~ & \textbf{CLIP} & \textbf{Inception} \\
        \midrule
        \textbf{AFHQ -Wild} & 0.0162 & 4.07E-192 \\ 
        \textbf{AFHQ -Dogs} & 0.1893 & 1.40E-31 \\
        \textbf{CelebA} & 0.0674 & 2.10E-20 \\
        \textbf{NotImageNet32} & 0.1328 & 0.0049 \\
        \textbf{ImageNet} & 0.093 & 6.34E-59 \\ \bottomrule
    \end{tabular}
    \label{tab:normality}
\end{table}

The results, reported in Table~\ref{tab:normality}, indicate that the Inception features are non-Gaussian and that across the board CLIP achieves better scores. Surprisingly, Inception features achieved the best score, by far, on our synthetic dataset and not on the original ImageNet dataset on which they were trained. We hypothesize that this is because our images were also generated by a deep neural network, albeit with a much different structure than the Inception network, and thus share certain characteristics as a result. 


\section{Conclusions}
To summarise, we generated a high-quality synthetic dataset and compared the standard empirical metrics such as FID and IS to probabilistic f-divergences such as KL and RKL. We first observe that the empirical metrics show good correlation, so they do capture important trends. However, they are much more volatile and not all significant gains in one of the metrics correspond to observable gains in one of the KL divergences. We also observed that IS and its IS$\infty$ extensions performed significantly worse compared to all other metrics. Finally, we investigated the standard use of the Inception features and show that, especially on benchmarks that are not ImageNet, they are outperformed by the more general-purpose CLIP features. 
 
Given these observations we recommend:
\begin{itemize}
    \item Drop the use of Inception Score, and used FID$_\infty$ instead of FID.
    \item Use multiple metrics (e.g. FID$_\infty$, KID and Clear FID) to try and control the volatility in scores.
    \item Replace the inception network with CLIP in FID. We made the code for FID based on CLIP available \footnote{https://github.com/eyalbetzalel/fcd}.
    \item Advocate NotImageNet32 as test-bed for generative models.
\end{itemize}

\bibliographystyle{unsrt}
\bibliography{main}

\newpage

\newpage
\appendix

\section{Qualitative analysis on Additional Datasets}

We perform anomaly detection base on the Based on the principles outlined in section 5.1 on AFHQ - Dogs and CelebA as well. As one can see, the results are consistent with those published in the article in section 5.

\begin{figure}[h]
    \centering
    \begin{minipage}{.5\textwidth}
    \centering
    \begin{center}
    \begin{subfigure}
        \centering
        \includegraphics[scale=0.3]{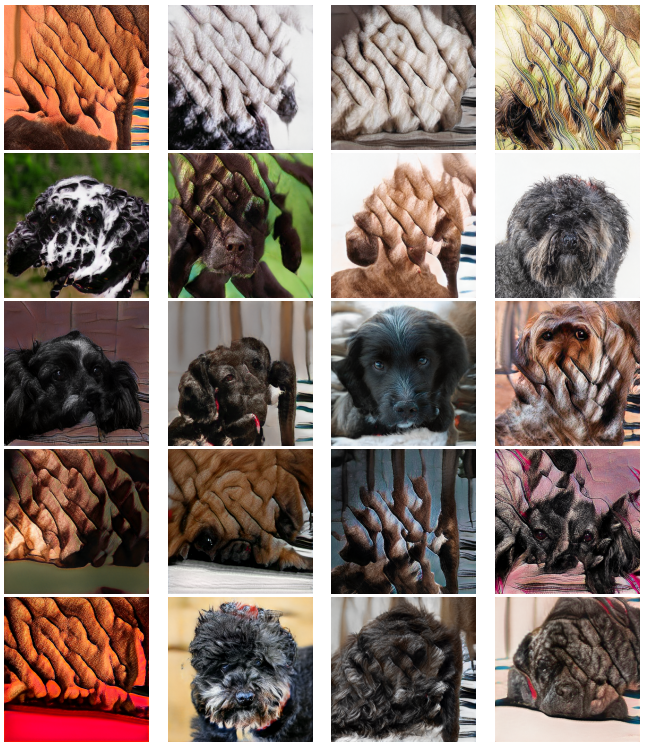}
        \caption{Inception - low probability (Dogs).}
        \label{fig:low_inception}
     \end{subfigure}
     \end{center}
     \end{minipage}%
\hfill
    \begin{minipage}{.5\textwidth}
    \centering
    \begin{center}
    \begin{subfigure}
        \centering
        \includegraphics[scale=0.3]{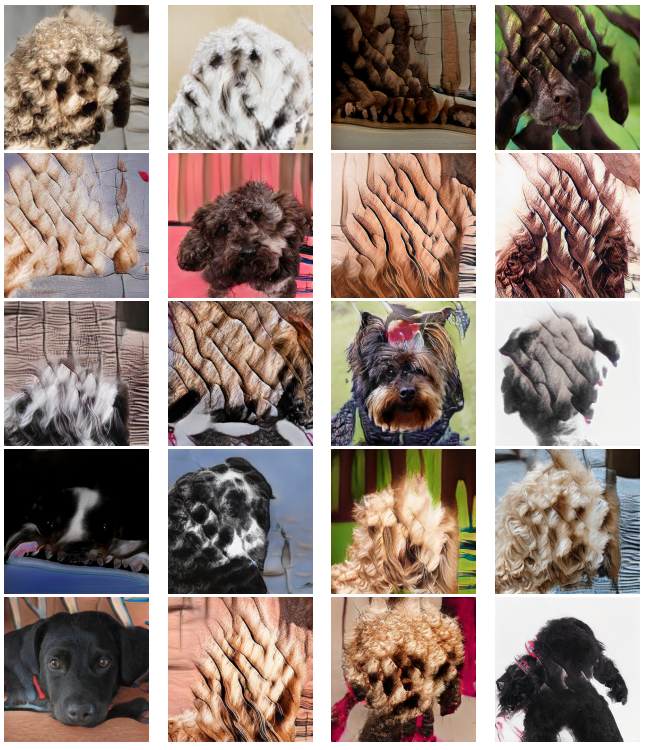}
        \caption{CLIP - low probability (Dogs).}
        \label{fig:low_clip}
     \end{subfigure}
     \end{center}
     \end{minipage}

\end{figure}

\begin{figure}[h]
    \centering
    \begin{minipage}{.5\textwidth}
    \centering
    \begin{center}
    \begin{subfigure}
        \centering
        \includegraphics[scale=0.3]{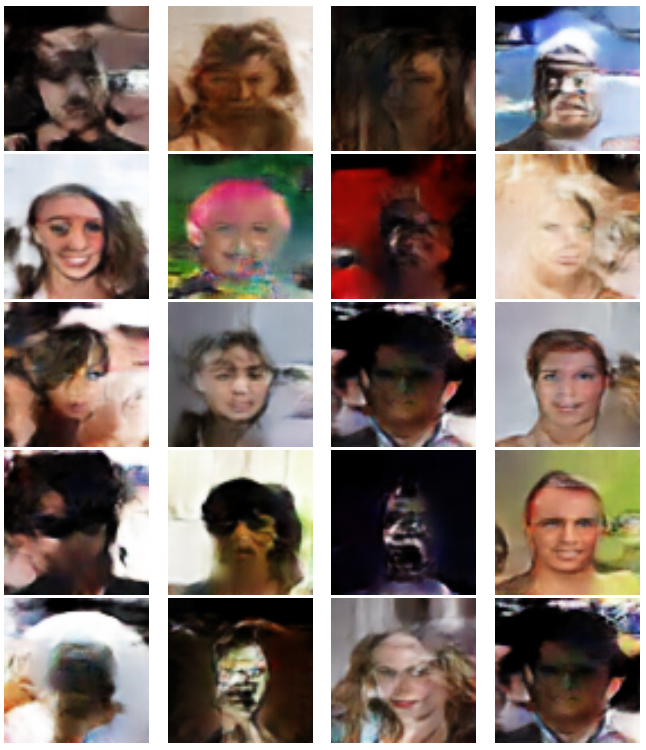}
        \caption{Inception - low probability (CelebA).}
        \label{fig:low_inception}
     \end{subfigure}
     \end{center}
     \end{minipage}%
\hfill
    \begin{minipage}{.5\textwidth}
    \centering
    \begin{center}
    \begin{subfigure}
        \centering
        \includegraphics[scale=0.3]{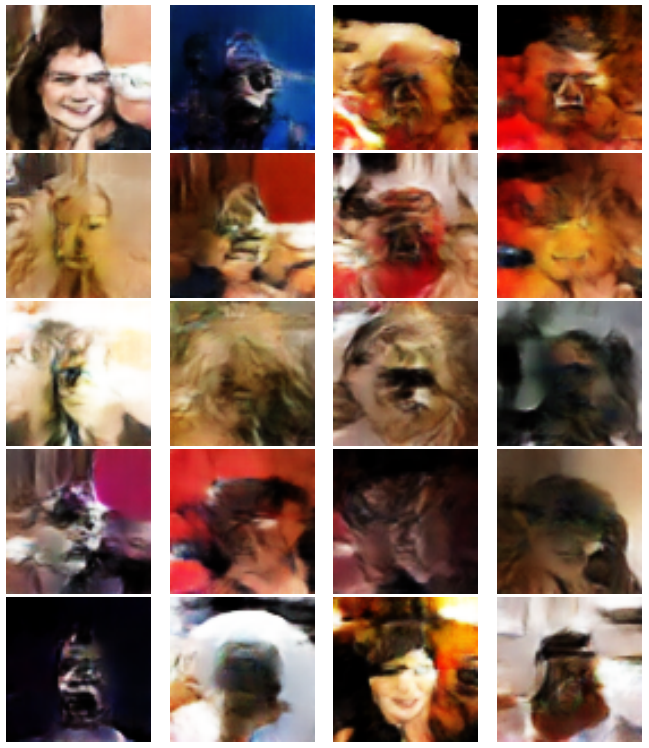}
        \caption{CLIP - low probability (CelebA).}
        \label{fig:low_clip}
     \end{subfigure}
     \end{center}
     \end{minipage}

\end{figure}

\newpage
\section{Volatility Analysis of High Quality Models}

\begin{figure*}[!h]
    \centering
    \captionsetup{justification=centering}
    \includegraphics[scale=0.3]{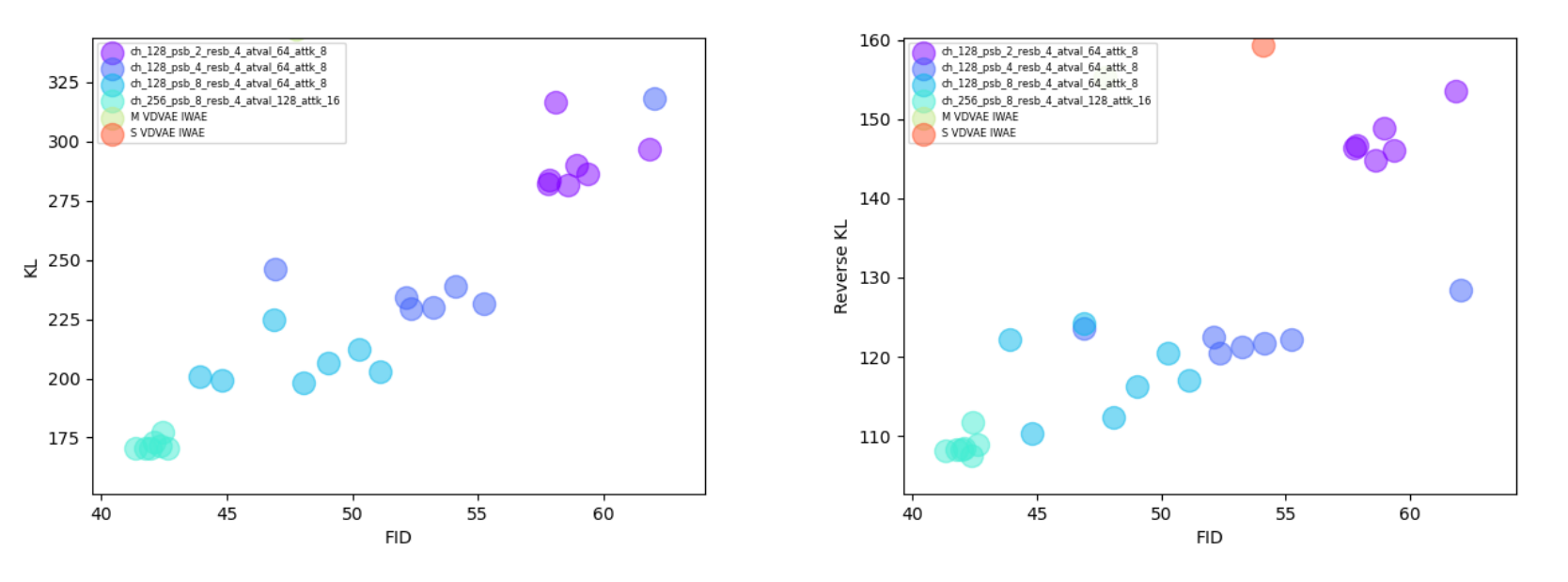}
    \caption {Evaluation metrics along the training of four pixelsnail and two VD-VAE models of varying sizes. Zoom in on high quality models.}
    \label{fig:scatter_zoom}
\end{figure*}


In Fig \ref{fig:scatter_zoom} one can see that FID score dramatically change although there is not much change in the KL or in the RKL metrics. This may indicate on the volatility of this method. 

\section{Technical details on experiment's generative models architecture}

As mentioned in section 4, we create different models by setting different hyper-parameters in order to compare performances between them. In order to enable accurate reproduction capability we describe the set of parameters we used.  

\subsection{PixelSnail}

The PixelSNAIL architecture is primarily composed of two main components: residual block, which applies several 2D-convolutions to its input, each with residual connections. The other is the attention block, which performs a single key-value lookup. It projects the input to a lower dimensionality to produce the keys and values and then uses softmax-attention. The model is built from serval PixelSnail blocks concat to one another, each interleaves the residual blocks and
attention blocks mentioned earlier. We used Adam optimizer with LR 0.0001 and MultiplicativeLR scheduler with lambada LR 0.999977. The loss function changed to the mean cross-entropy over 512 discrete clusters. All the other parameters that make up a model are described in table ~\ref{tab:pixelsnail_hyp}.

\begin{table}[hbt!]\label{tab:pixelsnail_hyp}
    \centering
    \caption{PixelSnail hyper-parameters}
    \begin{tabular}{c c c c c c}
    \toprule
       Size & Channels & PixelSnail blocks & Residual blocks & Attention values & Attention keys \\ \midrule
        S & 128 & 2 & 4 & 64 & 8  \\
        M & 128 & 4 & 4 & 64 & 8  \\
        L & 128 & 8 & 4 & 64 & 8  \\ 
        XL & 256 & 8 & 4 & 128 & 16  \\ 

        \bottomrule
    \end{tabular}
    \label{tab:pixelsnail_hyp}
\end{table}

\subsection{VD-VAE}

VD-VAE network is built from an encoder and decoder. In the encoder, there are regular blocks, which get an input and outputs output with the same dimension, and down-rate blocks that get input and output an output with a lower dimension. The difference between these two blocks is an avg\_pool2d at the end of the down-rate block. In the decoder, there are regular blocks and mixin blocks. the regular blocks get an input and outputs output with the same dimension. The input is fed from the previous layer and the parallel layer in the encoder. The mixin block performs interpolation to a higher dimension.

\begin{table}[hbt!]\label{tab:vdvae}
    \centering
    \caption{VD-VAE hyper-parameters}
    \begin{tabular}{c c c c c c}
    \toprule
       Size & Encoder & Decoder\\ \midrule
        S & 32x5, 32d2, 16x4, 16d2, 8x4, 8d2, 4x4, 4d4, 1x2 & 1x2, 4m1, 4x4, 8m4, 8x3, 16m8, 16x8, 32m16, 32x20 \\
        M & 32x10, 32d2, 16x5, 16d2, 8x8, 8d2, 4x6, 4d4, 1x4 & 1x2, 4m1, 4x4, 8m4, 8x8, 16m8, 16x10, 32m16, 32x30 \\

        \bottomrule
    \end{tabular}
    \label{tab:vdvae}
\end{table}

In table ~\ref{tab:vdvae} \textbf{x} means how many regular blocks are concatenated in a row. For example, 32x10 means 10 blocks in a row with a 32-channel input. \textbf{d} means a down-rate block. the number after tells the factor of the pooling. \textbf{m} means a unpool (mixin) block, for example, 32m16 means 32 is the output dimensionality with 16 layers in the mixin block. \\
Other hyper-parameters that were changed are EMA rate to 0.999, warm-up iterations to 1, learning rate to 0.00005, grad clip to 200, and skip threshold to 300.  We used Adam optimizer with $\beta_1=0.9$ and $\beta_2=0.9$. Other hyper-parameters configure as mentioned in VD-VAE article. 

\section{Supplementary models correlation measurements}

In table ~\ref{tab:pearson} one can see that Pearson correlation is high for most of evaluation methods. This fact is consistent with the conclusion presented in the article on the ability of current evaluation methods to capture trends. 

\begin{table}[hbt!]\label{tab:pearson}
    \centering
    \begin{tabular}{c c c c c c c c c}
    \hline
         & KL & RKL & FID & IS & IS$_\infty$ & KID & FID$_\infty$ & Clean FID \\ \hline
        KL & 1 & 0.976 & 0.8217 & 0.7088 & 0.5656 & 0.9011 & 0.911 & 0.8962 \\ \hline
        RKL & 0.976 & 1 & 0.7839 & 0.6559 & 0.5279 & 0.8552 & 0.8585 & 0.8493 \\ \hline\hline
        FID & 0.8217 & 0.7839 & 1 & 0.9441 & 0.9053 & 0.9771 & 0.9583 & 0.9829 \\ \hline
        IS & 0.7088 & 0.6559 & 0.9441 & 1 & 0.9657 & 0.9047 & 0.8858 & 0.9139 \\ \hline
        IS $\infty$ & 0.5656 & 0.5279 & 0.9053 & 0.9657 & 1 & 0.8301 & 0.799 & 0.8407 \\ \hline
        KID & 0.9011 & 0.8552 & 0.9771 & 0.9047 & 0.8301 & 1 & 0.9825 & 0.998 \\ \hline
        FID $\infty$ & 0.911 & 0.8585 & 0.9583 & 0.8858 & 0.799 & 0.9825 & 1 & 0.9863 \\ \hline
        Clean FID & 0.8962 & 0.8493 & 0.9829 & 0.9139 & 0.8407 & 0.998 & 0.9863 & 1 \\ \hline
    \end{tabular}
    \caption{Pearson's $\rho$ Correlation}
    \label{tab:pearson}
\end{table}

In table ~\ref{tab:spearman} we present Spearman ranking correlation, other ranking correlation method that is similar to Kandell's $\tau$ and presented similar results. 

\begin{table}[hbt!]\label{tab:spearman}
    \centering
    \begin{tabular}{c c c c c c c c c}
    \hline
         & KL & RKL & FID & IS & IS$_\infty$ & KID & FID$_\infty$ & Clean FID \\ \hline
        KL & 1 & 0.9779 & 0.8449 & 0.7394 & 0.6064 & 0.9201 & 0.9353 & 0.9242 \\ \hline
        RKL & 0.9779 & 1 & 0.8118 & 0.6921 & 0.5693 & 0.8828 & 0.8883 & 0.8865 \\ \hline\hline
        FID & 0.8449 & 0.8118 & 1 & 0.9238 & 0.8934 & 0.9587 & 0.9165 & 0.9627 \\ \hline
        IS & 0.7394 & 0.6921 & 0.9238 & 1 & 0.9548 & 0.8904 & 0.847 & 0.8799 \\ \hline
        IS$_\infty$ & 0.6064 & 0.5693 & 0.8934 & 0.9548 & 1 & 0.799 & 0.7422 & 0.7922 \\ \hline
        KID & 0.9201 & 0.8828 & 0.9587 & 0.8904 & 0.799 & 1 & 0.9656 & 0.9964 \\ \hline
        FID$_\infty$ & 0.9353 & 0.8883 & 0.9165 & 0.847 & 0.7422 & 0.9656 & 1 & 0.9715 \\ \hline
         Clean FID & 0.9242 & 0.8865 & 0.9627 & 0.8799 & 0.7922 & 0.9964 & 0.9715 & 1 \\ \hline
    \end{tabular}
    \caption{Spearman's $\rho$ Ranking Correlation}
    \label{tab:spearman}
\end{table}


\end{document}